\documentclass{article}
\usepackage{ijcai20}

\usepackage{times}

\usepackage{soul}
\usepackage{url}
\usepackage[hidelinks]{hyperref}
\usepackage[utf8]{inputenc}
\usepackage[small]{caption}
\usepackage{graphicx}
\usepackage{amsmath}
\usepackage{booktabs}
\usepackage{multirow}
\usepackage{color}
\usepackage{multirow}
\usepackage{makecell}
\usepackage{amssymb}
\usepackage{xspace}
\makeatletter
\DeclareRobustCommand\onedot{\futurelet\@let@token\@onedot}
\def\@onedot{\ifx\@let@token.\else.\null\fi\xspace}
 
\def\ie{\emph{i.e}\onedot}

\def\etal{\emph{et al}\onedot}
\makeatother

\title{Semi-Autoregressive Transformer for Image Captioning}

\author{
Yuanen Zhou$^1$  \and
Yong Zhang$^2$   \and
Zhenzhen Hu$^1$      \and
Meng Wang$^{1,3}$ \\
\affiliations
$^1$School of Computer Science and Information Engineering, Hefei University of Technology\\
$^2$Tencent AI Lab\\
$^3$Key Laboratory of Knowledge Engineering with Big Data (Ministry of Education)
\emails
\{y.e.zhou.hb,zhangyong201303,huzhen.ice,eric.mengwang\}@gmail.com}

\begin{document}

\maketitle

\begin{abstract}
	Current state-of-the-art image captioning models adopt autoregressive decoders, \ie they generate each word by conditioning on previously generated words, which leads to heavy latency during inference. To tackle this issue, non-autoregressive image captioning models have recently been proposed to significantly accelerate the speed of inference  by generating all words in parallel. However, these non-autoregressive models inevitably suffer from large generation quality degradation since they remove words dependence excessively. To make a better trade-off between speed and quality, we introduce a semi-autoregressive  model for image captioning~(dubbed as SATIC), which keeps the autoregressive property in global but generates words parallelly in local . Based on Transformer, there are only a few modifications needed to implement SATIC. Experimental results on the MSCOCO image captioning benchmark show that SATIC can achieve a good trade-off without bells and whistles. Code is available at {\color{magenta}\url{https://github.com/YuanEZhou/satic}}.
\end{abstract}

\section{Introduction}
Image captioning \cite{vinyals2015show,yang2019auto,pan2020x}, which aims at describing the visual content of an image with natural language sentence, is one of the important tasks to connect vision and language. 
Most  proposed models typically follow the encoder/decoder paradigm. In between,  convolutional neural network (CNN) is utilized to  encode an input image and recurrent neural networks (RNN) or Transformer \cite{vaswani2017attention} is adopted as  sentence decoder to generate a caption.
Current state-of-the-art models adopt autoregressive decoders which means that
they generate one word at each time step  by conditioning  on all previously produced words. 
Though impressive results have been achieved, these models suffer from high latency during inference  owing to the autoregressive property, which is  unaffordable for real-time industrial scenarios sometimes.

\begin{figure}[!t] 
	\centering
	\includegraphics[width=3.3in]{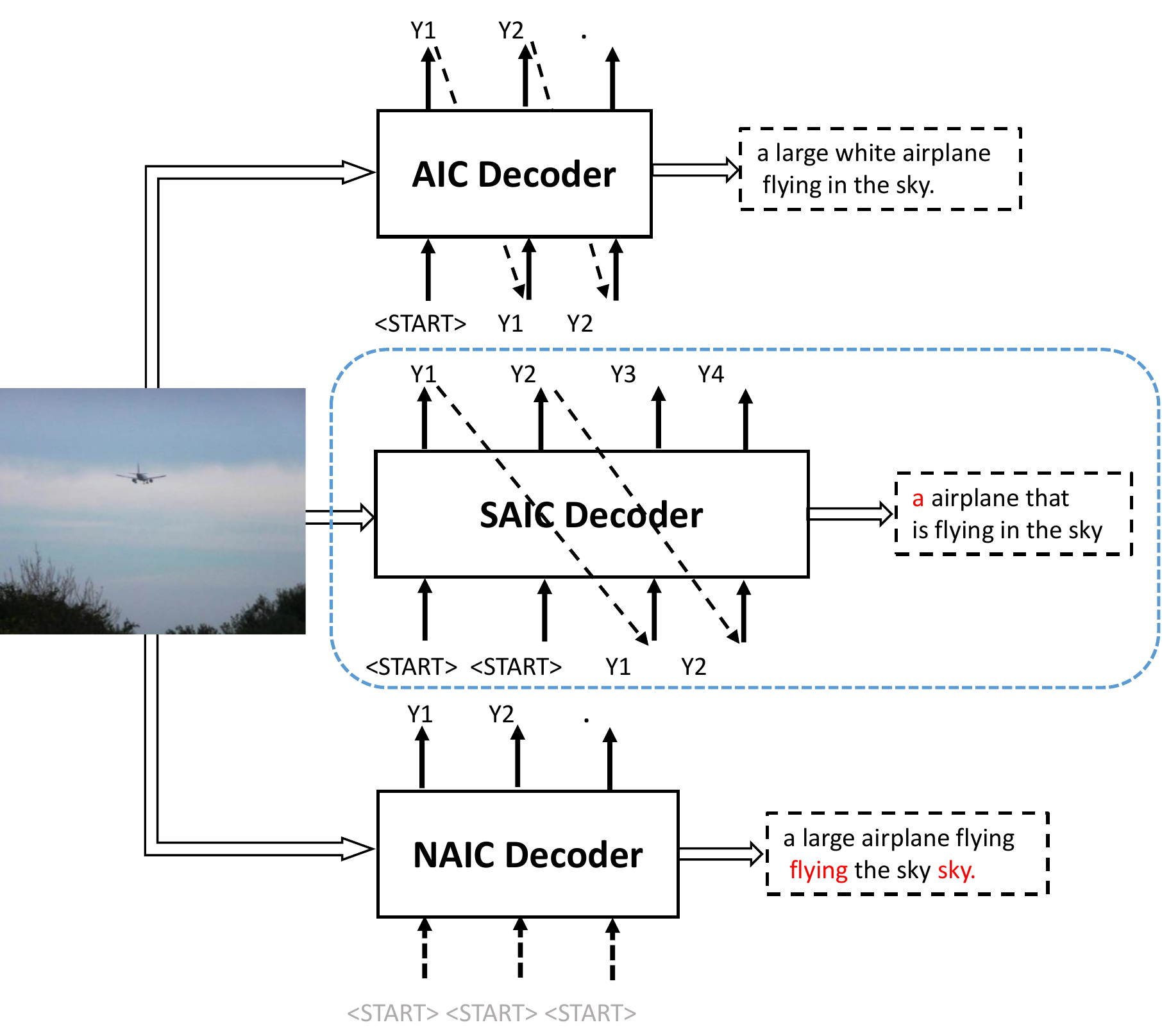}
	\caption{
		Given an image, autoregressive image captioning (AIC) model generates a caption word by word and Non-Autoregressive Image Captioning (NAIC) model outputs all words in parallel, while Semi-autoregressive image captioning (SAIC) model falls in between, which keeps the autoregressive property in global but outputs  words  parallelly in local.  We mark error words by red font.}
	\label{teaser}
\end{figure}

To tackle this issue,  there is an increasing interest to develop non-autoregressive decoding~\cite{gu2017non,lee2018deterministic,fei2019fast,guo2020non} to significantly accelerate inference speed  by generating all target words parallelly.  
These non-autoregressive models have basically the same structure as the autoregressive Transformer model \cite{vaswani2017attention}.
The difference lie in that non-autoregressive models generate all words independently~(as shown in the bottom of Figure~\ref{teaser}) instead of generating one word at each time step by conditioning on the previously produced words  as in autoregressive models~(as shown in the top of Figure~\ref{teaser}). 
However, these non-autoregressive models suffer from words repetition  or omission problem  compared to their autoregressive counterparts owing to removing the sequential dependence excessively.

To alleviate the above issue, some  methods have been proposed to seek a trade-off between speed and quality. For example, iteration refinement based methods~\cite{lee2018deterministic,gao2019masked,yang2021NACF} try to compensate for the word independence assumption by taking caption output from preceding iteration  as input and then polishing  it until reaching max iteration number or no change appears.  Nevertheless, it
needs multiple times refinement  to achieve better quality, which hurts decoding speed significantly.
Some  works~\cite{fei2019fast,guo2019non} try to enhance the decoder input by providing more target side context information, while they commonly incorporate  extra modules and thus extra computing overhead. Besides,  partially non-autoregressive models~\cite{pnaic,ran2020learning} are proposed by considering a sentence as a series of concatenated word groups. The groups are generated parallelly in global while each word in group is predicted from left to right. Though  better trade-off is achieved, the training paradigm of such model is somewhat tricky because it must need to incorporate curriculum learning-based training tasks of group length prediction and invalid group deletion~\cite{pnaic}.

In contrast, the model~(SATIC) introduced in this paper can achieve similar trade-off performance but without bells and whistles during training. Specifically, SATIC also considers a sentence as a series of concatenated word groups, as similar with~\cite{pnaic,ran2020learning}. However, all words in a group are predicted in parallel while the groups are generated from left to right, as shown in the middle of Figure~\ref{teaser}. This means that SATIC keeps the autoregressive property in global and the non-autoregressive property in local and thus gets the best of both world. 
In other words, SATIC can directly inherit the mature training paradigm of autoregressive captioning models and get the speedup benefit of non-autoregressive captioning models.

We evaluate SATIC model on the challenging MSCOCO \cite{chen2015microsoft} image captioning benchmark. 
Experimental results show that SATIC  achieves a better balance between speed, quality and  easy training. Specifically, SATIC generates captions better than non-autoregressive models and faster than  autoregressive models and is easy to be trained than partially non-autoregressive models~\cite{pnaic}. 
Besides, we conduct substantial ablation studies to better understand  the effect of each component of the whole model.

\section{Related Work}
\paragraph{Image Captioning.} 
Over the last few years, a broad collection of methods have been proposed in the field of image captioning. In a nutshell, we have gone through grid-feature~\cite{xu2015show,jiang2020defense} then region-feature~\cite{anderson2018bottom} and relation-aware visual feature~\cite{yao2018exploring,yang2019auto} on the image encoding side. On the sentence decoding side, we have witnessed LSTM~\cite{vinyals2015show}, CNN~\cite{gu2017empirical} and Transformer~\cite{cornia2020meshed} equipped with various attention~\cite{huang2019attention,zhou2020more,pan2020x} as decoder. On the training side, models are typically trained by step-wise cross-entropy loss and then Reinforcement Learning~\cite{rennie2017self}, which enables the use of non-differentiable caption metrics as optimization objectives and makes a notable achievement. 
Recently, vision-language pre-training has also been adopted for image captioning and show impressive result. These models~\cite{zhou2020unified,li2020oscar} are firstly pretrained on large image-text corpus and then finetuned.
It is noteworthy that it's not fair to directly compare them with non-pretraining-finetuning methods.
Though impressive performance has been achieved, most state-of-the-art models adopt autoregressive decoders and thus suffer from high latency during inference.

\paragraph{Non-Autoregressive Decoding.}
Due to the downside of autoregressive decoding, Non-autoregressive decoding has firstly aroused  widespread attention in the community of Neural Machine Translation~(NMT). 
Non-autoregressive NMT was first proposed in~\cite{gu2017non} to significantly improve the inference speed by generating all target-side words in parallel. While the decoding speed is improved, it often suffers from word repetition or omission problem due to removing words dependence excessively.  Some methods have been proposed to overcome this problem, including knowledge distillation \cite{gu2017non}, well-designed decoder input \cite{guo2019non}, auxiliary regularization terms \cite{wang2019non},
iterative refinement \cite{lee2018deterministic}, and partially-autoregressive model~\cite{ran2020learning,wang2018semi}.
Following similar research roadmap, non-autoregressive decoding has recently been introduced to visual captioning task~\cite{gao2019masked,guo2020non,fei2019fast,yang2021NACF,pnaic}. This work pursues the semi-autoregressive decoding  in NMT~\cite{wang2018semi} for image captioning and further explores its effectiveness under the context of reinforcement training.

\section{Background}
\subsection{Autoregressive Image Captioning}
Given an image $I$ and the associated target sentence $y=(y_1,..., y_T)$, AIC models 
the conditional probability as: 
\begin{equation}
p(y | I; \theta)=\prod_{i=1}^{T} p\left(y_{i} | y_{<i}, I; \theta \right),
\label{aic}
\end{equation}
where $\theta$ is the model's parameters and $y_{<i}$ represents the words sequence before the $i$-th word of target $y$. 
During inference, the sentence is generated word by word sequentially, which causes high inference latency. 

\subsection{Non-Autoregressive Image Captioning}
Non-autoregressive image captioning~(NAIC) models are recently proposed to accelerate the decoding  speed by discarding the words dependence within the  sentence. 
A NAIC model generates all words simultaneously and the conditional probability can be modeled as:
\begin{equation}
p(y | I; \theta)=\prod_{i=1}^{T} p\left(y_{i} | I; \theta \right). 
\label{naic}
\end{equation}
During inference, all words are parallelly generated in one pass and thus the inference speed is significantly improved. 
However, these NAIC models inevitably suffer from large generation quality degradation since they remove words dependence excessively. 

\begin{figure*}[!t] 
	\centering
	\resizebox{0.7\textwidth}{!}{
		\includegraphics[width=10in]{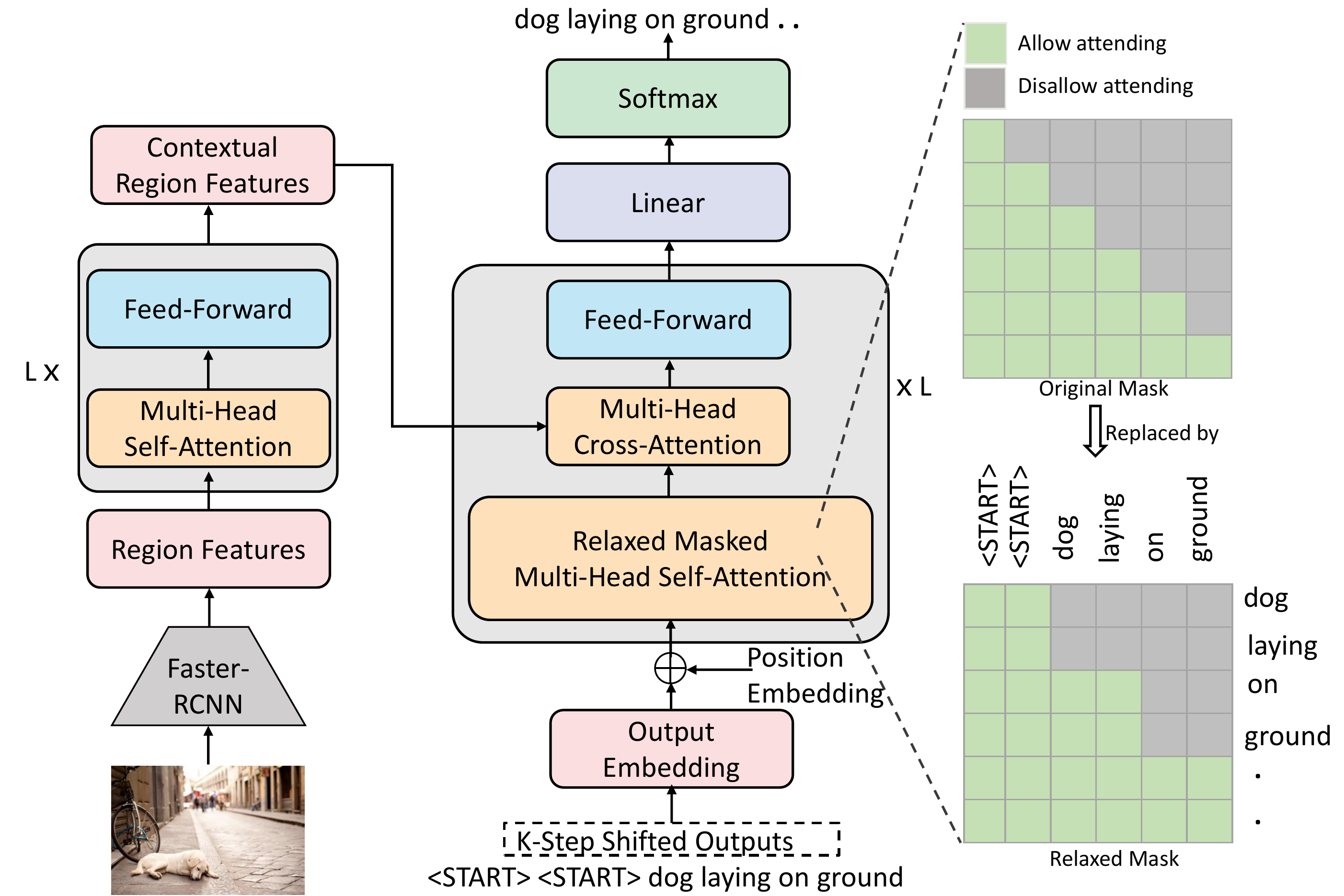}
	}
	\caption{
		Illustration of  Transformer-based semi-autoregressive image captioning model~(SATIC), which composes of an encoder and a decoder. 
		Without loss of generality, we set K=2 for convenience. Notice that the Residual Connections, Layer Normalization are omitted. 
	}
	\label{framework}
\end{figure*}

\section{Approach} 
In this section, we first present the architecture of our SATIC model built on the well-known Transformer~\cite{vaswani2017attention}
and then introduce the training procedure  for model optimization. 

\subsection{Transformer-Based SATIC Model} 
Given the image region features  extracted by a pre-trained Faster-RCNN model~\cite{ren2015faster,anderson2018bottom}, 
SATIC aims to generate a caption in a semi-autoregressive manner. 
The architecture of SATIC model is shown in Figure~\ref{framework},  
which consists of an encoder and decoder.

\paragraph{Image Features Encoder.} 
The encoder takes the image region features as inputs and outputs the contextual region features. It consists of a stack of $L$ identical layers.  Each layer has two sublayers. The first is  a multi-head self-attention sublayer and the second is a position-wise feed-forward sublayer. Both sublayers are followed by residual connection~\cite{he2016deep} and layer normalization~\cite{ba2016layer} operations for stable training.
Multi-head self-attention builds on scaled dot-product attention, which operates on a query Q, key K and value V as:
\begin{equation}
    Attention(Q,K,V) = softmax(\frac{QK^T}{\sqrt{d_k}})V,
\end{equation}
where $d_k$ is the dimension of the key.  
Multi-head self-attention  firstly  projects the   queries, keys and values $h$ times with different learned linear projections and then computes scaled dot-product attention for each one. After that, it concatenates the results and projects them with another learned linear projection:
\begin{gather}
    H_i = Attention(QW_i^Q, KW_i^K, VW_i^V), \\
    MultiHead(Q,K,V) = Concat(H_1,\dots H_h)W^O ,
\end{gather}
where $W_i^Q, W_i^K \in \mathbb{R}^{d_{model}\times d_k}$, $W_i^V \in \mathbb{R}^{d_{model}\times d_v}$ and $W^O \in \mathbb{R}^{hd_v \times d_{model}}$ . The self-attention in the encoder performs attention over itself, i.e.,  ($Q=K=V$), which is the image region features in the first layer.
After a multi-head self-attention sublayer,  the position-wise feed-forward sublayer~(FFN) is applied to each position separately and identically:
\begin{equation}
    FFN(x) = max(0, xW_1+b_1)W_2+b_2,
\end{equation}
where $W_1\in \mathbb{R}^{d_{model}\times d_{ff}}$, $W_2\in \mathbb{R}^{d_{ff}\times d_{model}}$, $b_1\in \mathbb{R}^{d_{ff}}$ and $b_2\in\mathbb{R}^{d_{model}}$ are learnable parameter matrices.

\paragraph{Captioning Decoder.} 
The decoder takes contextual region features and previous word embedding features as input  and outputs predicted words probability. To make use of order information, position encodings~\cite{vaswani2017attention} are added to word embedding features. 
Different from original autoregressive Transformer model, which outputs a  word at each step, SATIC takes a group of words as input and outputs a group of words  at each step during decoding. 
Each group contains $K$ consecutive words. At the beginning of decoding, we feed the model with $K$ $<$START$>$ symbols to predicate $y_1,...,y_K$ and then $y_1,...,y_K$ are fed as input to predicate $y_{K+1},...,y_{2K}$ in parallel. This process will continue until the end of sentence. An intuitive example is shown in the middle of Figure~\ref{teaser} with $K=2$.
The decoder also consists of $L$ identical layers and each layer has three sublayers:  a relaxed  masked multi-head self-attention sublayer, multi-head cross-attention sublayer and a position-wise feed-forward sublayer. Residual connection and layer normalization are also applied after each sublayer. The multi-head cross-attention is similar with the multi-head self-attention mentioned above except that the key and value are now contextual region features and the query is the output of its last sublayer. 
It is particularly  worth emphasizing that original masked multi-head attention  is replaced with the relaxed  masked multi-head self-attention. The only difference between the two sublayer lie in the self-attention mask. Specifically, the original lower triangular matrix mask is now replaced by the relaxed  mask. Formally, given the caption length $T$ and group size $K$, the relaxed  mask $M \in \mathbb{R}^{T \times T}$ is defined as:
\begin{gather}
     M[i][j]=
\begin{cases}
0 & \text{if $j < ([(i-1)/K]+1)\times K$, allow} \\
-\infty & \text{other, \quad   disallow attending}
\end{cases},
\end{gather}
where $i,j \in [1,T]$ and $[\cdot]$ denotes floor operation. An intuitive example is shown in the right of Figure~\ref{framework}, where $T=6$ and $K=2$.
As a consequence, the scaled dot-product attention in relaxed  masked multi-head self-attention module is modified to:
\begin{equation}
    Attention(Q,K,V) = softmax(\frac{QK^T}{\sqrt{d_k}}+M)V.
\end{equation}
.
\subsection{Training} 
Since SATIC model keeps the autoregressive property in global and the non-autoregressive property in local, it  gets the best of both world and the conditional probability can be  formulated as:
\begin{equation}
p(y | I; \theta)=\prod_{t=1}^{[(T-1)/K]+1} p\left(G_{t} |G_{<t}, I; \theta \right),
\label{satic}
\end{equation}
where $G_{<t}$ represents  the groups  before $t$-th group and $G_{t}=y_{(t-1)K+1},...,y_{tK}$ except for the last group which may have less than $K$ words. If the length of padded word sequence can't be divided by $K$, remaining words only keep in output but not in input.
Comparing Equation~\ref{aic} and Equation~\ref{naic}  with Equation~\ref{satic}, we can find that autoregressive and non-autoregressive image captioning models are special cases of SATIC model when $K=1$ and $K=T$ respectively.  
At the first training stage, we optimize the model by minimizing  cross-entropy loss:
\begin{equation}
\begin{aligned}
L_{XE}(\theta)=&-\sum_{t=1}^{[(T-1)/K]+1}\log p\left(G_{t} |G_{<t}, I; \theta \right)\\
      =&-\sum_{t=1}^{[(T-1)/K]+1}\sum_{i=(t-1)K+1}^{tK}\log p\left(y_{i} |G_{<t}, I; \theta \right).
\end{aligned}
\label{eqn:sat}
\end{equation}
At the second training stage, we finetune the model using  self-critical training~\cite{rennie2017self} and the gradient can be expressed as:
\begin{equation}
    \nabla_\theta L_{SC}(\theta)=-\frac{1}{N}\sum_{n=1}^N(R(\hat{y}^n_{1:T})-b)\nabla_\theta\log p(\hat{y}^n_{1:T}|I;  \theta),
\end{equation}
where $R$ is the CIDEr~\cite{Vedantam2015CIDEr} score function, and $b$ is the baseline score. We adopt the baseline score proposed in~\cite{luo2020better}, where 
the baseline score is defined as the average reward of the rest samples rather than original greedy decoding reward. We sample $N=5$ captions for each image and $\hat{y}_{1:T}^n$ is the $n$-th sampled caption.

\begin{table*}[htb]
	\centering
	\resizebox{\textwidth}{!}{
		\begin{tabular}{l|cccccc|cc}
			\toprule
			Models & \multicolumn{1}{c}{BLEU-1} & \multicolumn{1}{c}{BLEU-4} & \multicolumn{1}{c}{METEOR} & \multicolumn{1}{c}{ROUGE} & \multicolumn{1}{c}{SPICE} & \multicolumn{1}{c}{CIDEr} & \multicolumn{1}{|c}{Latency} & \multicolumn{1}{c}{Speedup} \\
			\midrule
			\midrule
			\multicolumn{9}{l}{\textbf{Autoregressive models}} \\
			\midrule
			NIC-v2 \cite{vinyals2015show} & / & 32.1  & 25.7  & / & / & 99.8  & / & / \\
			Up-Down \cite{anderson2018bottom} & 79.8  & 36.3  & 27.7  & 56.9  & 21.4  & 120.1 & / & / \\
			AOANet \cite{huang2019attention} & 80.2 & 38.9 & 29.2 & \bf58.8&  22.4 & 129.8 & / & / \\
			M2-T$^\dag$ \cite{cornia2020meshed} & \bf80.8 & \bf39.1 & \bf29.2 & 58.6 & 22.6 & \bf131.2  & / & / \\
			AIC$^\dag$ ($\text{bw}=1$) &  80.5&	38.8&	29.0 &	58.7	& 22.7&	128.0  &   135ms    &  \bf2.25$\times$ \\
			AIC$^\dag$ ($\text{bw}=3$) &  \bf80.8&	\bf39.1 &	29.1 &	\bf58.8	& \bf22.9&	129.7  &   304ms    &  1.00$\times$ \\
			\midrule 
			\multicolumn{9}{l}{\textbf{Non-autoregressive models}} \\
			\midrule
			MNIC$^\dag$ \cite{gao2019masked} & 75.4  & 30.9  & 27.5  & 55.6  & 21.0    & 108.1 & - & 2.80$\times$ \\
			FNIC $^\dag$\cite{fei2019fast} 						  & /     & 36.2  & 27.1  & 55.3  &20.2     &115.7  & -   &  8.15$\times$ \\
			MIR $^\dag$\cite{lee2018deterministic} 						  & /     & 32.5  & 27.2  & 55.4  &20.6     &109.5  & -   &  1.56$\times$ \\
			CMAL $^\dag$\cite{guo2020non} 						  & \bf80.3     & \bf37.3  & \bf28.1  & \bf58.0  &\bf21.8     & \bf124.0  & -   &  \bf13.90$\times$ \\
			\midrule
			\multicolumn{9}{l}{\textbf{Partially Non-autoregressive models  }} \\
			\midrule
			PNAIC(K=2) $^\dag$\cite{pnaic} 						  & 80.4     & 38.3  & \bf29.0  & 58.4  &22.2     & \bf129.4  & -   &  2.17$\times$ \\
			PNAIC(K=5) $^\dag$\cite{pnaic} 						  & 80.3     & 38.1  & 28.7  & 58.3  &22.0     & 128.5  & -   &  3.59$\times$ \\
			PNAIC(K=10) $^\dag$\cite{pnaic} 						  & 79.9     & 37.5  & 28.2  & 58.0  &21.8     & 125.2  & -   &  5.43$\times$ \\	 \hline
			SATIC(K=2, \text{bw}=3) $^\dag$						  & \bf80.8     & \bf38.4  & 28.8  & 58.5  &\bf22.7     & 129.0  & 184ms   &  1.65$\times$ \\
			SATIC(K=2, \text{bw}=1) $^\dag$						  & 80.7     & 38.3  & 28.8  & \bf58.5  &22.7     & 128.8   & 76ms   &  4.0$\times$ \\	 
			SATIC(K=4, \text{bw}=3) $^\dag$						  & 80.7     & 38.1  & 28.6  & 58.4  &22.4     & 127.4   & 127ms   &  2.39$\times$ \\	 
			SATIC(K=4, \text{bw}=1) $^\dag$						  & 80.6     & 37.9  & 28.6  & 58.3  &22.3     & 127.2   & 46ms   &  6.61$\times$ \\
			SATIC(K=6, \text{bw}=3) $^\dag$						  & 80.6     & 37.6  & 28.3  & 58.1  &22.1     & 126.2   & 119ms   &  2.55$\times$ \\
			SATIC(K=6, \text{bw}=1) $^\dag$						  & 80.6     & 37.6  & 28.3  & 58.1  &22.2     & 126.2   & 35ms   &  \bf8.69$\times$ \\

			\bottomrule
		\end{tabular}%
	}
	\caption{Performance comparisons with different evaluation metrics on the MS COCO offline test set. 
		``$\dag$" indicates the model is based on Transformer architecture. 
		AIC is our implementation of the Transformer-based autoregressive model, which has the same structure as SATIC models and is used as the teacher model for sequence knowledge distillation. 
		``/" denotes that the results are not reported. 
		``bw" denotes the beam width used for beam search. 
		Latency is the time to decode a single image without minibatching, 
		averaged over the whole test split. 	
		The Speedup values  are from the corresponding papers. Since  latency is influenced by platform, implementation and hardware, it is not fair to directly compare them.  A fairer alternative way is to compare speedup, which is calculated based on their own baseline.
	}
	\label{offline}%
\end{table*}%

\section{Experiments}
\subsection{Dataset and Evaluation Metrics.} MSCOCO \cite{chen2015microsoft} 
is the widely used  benchmark for image captioning. 
We use the `Karpathy' splits \cite{karpathy2015deep} for offline experiments. This split contains $113,287$
training images, $5,000$ validation images and $5,000$ testing images. Each image has $5$ captions.
We also upload generated captions of MSCOCO
official testing set, which contains $40,775$ images for online evaluation.
We evaluate the quality of captions by standard  metrics\cite{chen2015microsoft} , including BLEU-1/4, METEOR, ROUGE, SPICE, and CIDEr, 
denoted as B1/4, M, R, S, and C, respectively.

\subsection{Implementation Details.}
Each image is represented as $10$ to $100$ region features with $2,048$ dimensions extracted by Anderson \etal~\cite{anderson2018bottom}. The dictionary is built by dropping the words that occur less than $5$ times and ends up with a vocabulary of $9,487$. Captions longer than $16$ words are truncated. 
Both our SATIC model and  autoregressive image captioning  base model~(AIC) almost follow the same model hyper-parameters setting as in  \cite{vaswani2017attention}
~($d_{model} = 512,d_k = d_v = 64,d_{ff} = 2048,L = 6,h = 8,p_{dropout} = 0.1$).
As for the training process,
we train AIC under cross entropy loss for 15 epochs with a mini
batch size of 10, and  optimizer in ~\cite{vaswani2017attention} is used with a
learning rate initialized by 5e-4 and the warmup step is set to $20000$. We increase the scheduled sampling probability by 0.05 every 5 epochs.
We then optimize the CIDEr score
with self-critical training for another 25 epochs with an initial learning
rate of 1e-5.
Our best SATIC model shares the same training script with AIC model except that we first initialize the weights~(weight-init) of SATIC model with the pre-trained AIC  model and replace the ground truth captions in the training set with sequence knowledge~(SeqKD)~\cite{kim2016sequence,gu2017non} results of AIC model with beam size $5$ during cross entropy training stage.
Unless otherwise indicate, latency represents the time to decode a single image without batching  averaged over the whole test split, and is tested on a
NVIDIA Tesla T4 GPU. The time for image feature extraction is not included in latency.

\begin{table*}[htb]
	\centering
	\resizebox{\textwidth}{!}{
		\begin{tabular}{@{\extracolsep{3pt}}@{\kern\tabcolsep}lrrrrrrrrrrrrrr}
			\toprule
			\multicolumn{1}{c}{\multirow{2}[4]{*}{Model}}  & \multicolumn{2}{c}{BLEU-1} & \multicolumn{2}{c}{BLEU-2} & \multicolumn{2}{c}{BLEU-3} & \multicolumn{2}{c}{BLEU-4} & \multicolumn{2}{c}{METEOR} & \multicolumn{2}{c}{ROUGE-L} & \multicolumn{2}{c}{CIDEr-D} \\
			\cmidrule{2-3}  \cmidrule{4-5} \cmidrule{6-7} \cmidrule{8-9} \cmidrule{10-11} \cmidrule{12-13} \cmidrule{14-15} %
			& \multicolumn{1}{c}{c5} & \multicolumn{1}{c}{c40} & \multicolumn{1}{c}{c5} & \multicolumn{1}{c}{c40} & \multicolumn{1}{c}{c5} & \multicolumn{1}{c}{c40} & \multicolumn{1}{c}{c5} & \multicolumn{1}{c}{c40} & \multicolumn{1}{c}{c5} & \multicolumn{1}{c}{c40} & \multicolumn{1}{c}{c5} & \multicolumn{1}{c}{c40} & \multicolumn{1}{c}{c5} & \multicolumn{1}{c}{c40} \\
			\midrule
			Up-Down$^*$ \cite{anderson2018bottom} & 80.2  & 95.2  & 64.1  & 88.8  & 49.1  & 79.4  & 36.9  & 68.5  & 27.6  & 36.7  & 57.1  & 72.4  & 117.9  & 120.5  \\
			AOANet \cite{huang2019attention} & 81.0  & 95.0  & 65.8  & 89.6  & 51.4  & 81.3  & 39.4  & 71.2  & 29.1  & 38.5  & 58.9  & 74.5  & 126.9  & 129.6  \\
			M2-T$^*$ \cite{cornia2020meshed} & 81.6& 96.0 &66.4 &90.8& 51.8& 82.7& 39.7& 72.8& 29.4 &39.0& 59.2 &74.8& 129.3& 132.1 \\
			\midrule
			CMAL~\cite{guo2020non} & 79.8 & 94.3 & 63.8 & 87.2 & 48.8 & 77.2 & 36.8 & 66.1 & 27.9 & 36.4 & 57.6 & 72.0 & 119.3 &  121.2 \\
			PNAIC~\cite{pnaic} & 80.1 & 94.4 & 64.0 & 88.1 & 49.2 & 78.5 & 36.9 & 68.2 & 27.8 & 36.4 & 57.6 & 72.2 & 121.6 &  122.0 \\
			SATIC~(K=4) & 80.3 & 94.5 & 64.4 & 87.9 & 49.2 & 78.2 & 37.0 & 67.2 & 28.2 & 37.0 & 57.8 & 72.6& 121.5 & 124.1 \\
			\bottomrule
		\end{tabular}%
	}
	\caption{Results on the online MSCOCO test server. $*$ denotes ensemble model.  }
	\label{online}%
\end{table*}%

\subsection{Quantitative Results}
In this section, we will analyse SATIC in detail by answering following questions.

\paragraph{How does SATIC perform compared with state-of-the-art models?} 
We  compare the performance of SATIC with other state-of-the-art autoregressive models, non-autoregressive models and a partially non-autoregressive model. 
Among the autoregressive models, M2-T and AIC are based on similar Transformer architecture, while others are based on LSTM \cite{hochreiter1997long}. 
For non-autoregressive models, MNIC and MIR adopts an iterative refinement strategy,  FNIC orders words detected in the image with an RNN and CMAL optimizes sentence-level reward.
PNAIC is a partially non-autoregressive model which divides sentence into word groups and the
groups are generated parallelly in global while each word
in group is predicted from left to right. 
As shown in Table~\ref{offline}, SATIC achieves comparable performance with state-of-the-art autoregressive models but with significant speedup. When $K=2$, SATIC achieves about $1.5 \times$ speedup while the caption quality  only degrades slightly compared with its autoregressive counterpart AIC model.
Compared with non-autoregressive models , SATIC obviously achieves a better trade-off between quality and speed by   outperforming  all the non-autoregressive models except CMAL in speedup metric.
Compared with most similar partially non-autoregressive model PNAIC, SATIC achieves similar speedup and caption evaluation results. It is worth nothing that SATIC outperforms PNAIC on SPICE metric, which concerns more on semantic propositional content. What's more, the training of SATIC is more easy and straightforward than PNAIC.
Overall, SATIC achieves a better trade-off between quality, speed and easy training.
We also show the results of online MSCOCO evaluation in Table~\ref{online}.

\paragraph{What is the effect of group size $K$?}
We test three different setting of the group size, \ie $K \in \{2,4,6\}$. From the bottom of Table~\ref{offline}, we can observe that a larger $K$ brings more significant speedup while the caption quality degrades moderately. For example, the decoding speed increases about $1.5 \times$ while the CIDEr score  drops about $1.5$ when $K$ grows from $2$ to $4$, and drops no more than $3$ when $K$ grows further to $6$.
This is intuitive since $K$ is the indicator of parallelization and also indicates that SATIC model is relatively stable to $K$ and has a good trade-off between speed and quality.

\paragraph{Can SATIC benefit from beam search?}
From the bottom of Table~\ref{offline}, we can also find that SATIC  benefits little~(CIDEr score only increases $0.2$) from beam search compared with its autoregressive counterpart AIC~(CIDEr score increases $1.7$) after self-critical training. There are two possibilities, the one is self-critical training make its output probability  concentrated and the other one is SATIC can not benefit from beam search.
To investigate whether SATIC can benefit from beam search, we test its output just after cross entropy training with different beam search width. From Table~\ref{beam}, we can observe that SATIC model can still benefit from beam search and there are two interesting phenomena: 1) SATIC with too large $K$ benefits less from beam search and 2) the effect of beam search is larger when without weight initialization and sequence knowledge distillation.
A  plausible explanation is that long-distance dependence is hard to capture and sequence knowledge distillation decreases dependence among words.
\begin{table}[htbp]
	\centering
	\resizebox{0.48\textwidth}{!}{
		\begin{tabular}{cc|rrrrrr}
			\toprule
			Models & bw & \multicolumn{1}{c}{B1} & \multicolumn{1}{c}{B4} & \multicolumn{1}{c}{M} & \multicolumn{1}{c}{R} & \multicolumn{1}{c}{S} & \multicolumn{1}{c}{C} \\
			\midrule
			\midrule
			\multicolumn{1}{l}{\textbf{w/ weight-init and SeqKD:}} \\
			\midrule
			\multirow{2}[0]{*}{K=2} & 1 & 79.3 & 36.2 &	28.2 &	57.4	&22.1& 	121.5 \\
			& 3 &  80.0 	&37.3 &	28.4 &	57.8& 	22.3 &	123.9  \\
			\midrule
			\multirow{2}[0]{*}{K=4} & 1 & 77.3 & 32.9 &	27.0 &	56.0	&20.5& 	111.0 \\
			& 3& 78.0 &	34.4 &	27.2 &	56.5	&20.9& 	114.5 \\
			\midrule
			\multirow{2}[0]{*}{K=6} & 1 & 77.3 & 33.3 &	26.7 &	56.0	&20.4& 	110.3 \\
			& 3 & 77.2 &	33.7 &	26.6 &	56.1	& 20.4& 	110.7 \\
			\midrule
			\multicolumn{1}{l}{\textbf{w/o weight-init and SeqKD :}} \\
			\midrule
			\multirow{2}[0]{*}{K=2} & 1 & 74.2 & 29.1 &25.8 &53.8	&19.5& 100.0 \\
			& 3 & 76.0 & 32.8 &26.8 &55.2	&20.6& 107.2 \\
			\midrule
			\multirow{2}[0]{*}{K=4} & 1 & 65.8 & 17.0 &21.8 &48.3	&16.1& 73.3 \\
			& 3& 69.7 & 20.9 &22.7 &50.2	&16.6& 79.6 \\
			\midrule
			\multirow{2}[0]{*}{K=6} & 1 & 67.6 & 17.6 &21.3 &48.6	&15.2& 73.5 \\
			& 3 & 68.8 & 19.1 &21.6 &49.6	&15.5& 76.3 \\
			\midrule
			\bottomrule
		\end{tabular}%
	}
	\caption{The results after XE training when using different beam search width.}
	\label{beam}%
\end{table}%

\paragraph{What is the effect of sequence knowledge distillation and weight initialization?}
We further investigate the effect of sequence knowledge distillation and weight initialization and results are shown in Table~\ref{tricks}. We can find that sequence knowledge distillation plays an important role in both XE and SC training stages and the effect is more significant in XE stage . Basically, the larger the $K$ is, the more obvious the effect is. This is intuitive since the ability of SATIC model to capture conditional probability  is undermined when $K$ grows and sequence knowledge distillation compensates for this by   reducing the complexity of data sets. SC can also alleviate sentence-level inconsistency by providing sentence-level reward.  
In addition to accelerate convergence, we can find that weight initialization  slightly improves the caption quality when $K$ is small but has important impact when $K$ is large.

\paragraph{What is the effect of batch size on latency?}
Above latency is measured with batch size set to $1$. However, there may be multiple requests at once in real application. So, we further investigate the latency under various batch size setting. From Table~\ref{latency}, we can find that SATIC can basically accelerate decoding even under large batch size.  We can also observe that the speedup is decline as batch size increases. This indicates that non-gpu program becomes a bottleneck when the runtime of gpu program is negligible.

\begin{table}[htbp]
\centering
\resizebox{0.48\textwidth}{!}{
\begin{tabular}{lccccc|ccccc}
\toprule
\multirow{2}{*}{Models} & \multicolumn{5}{c|}{XE} & \multicolumn{5}{c}{SC} \\ \cmidrule{2-11}
                &B1     &B4        & M       & S       & C      &B1        &B4     & M       & S       & C      \\ \midrule
K=2: \\
\midrule
Base            &74.2       &29.1          &25.8      &19.5       &100.0         &80.3         &37.6         &28.4         &22.0         &123.7         \\ 
+SeqKD          &78.8       &36.0        &28.0       &21.8        &120.4         & 80.5        &  38.4       &  28.7       &  22.6       &   128.1      \\ 
+Weight-init     &79.3      &36.2        &28.2         &22.1       &121.5         &80.7         &38.3         &28.8         &22.7         &128.8         \\ \midrule
K=4: \\
\midrule
Base            &65.8       &17.0          &21.8      &16.1       &73.3         &79.8         & 35.3        &27.3         &20.9         &119.5         \\ 
+SeqKD          &69.5       &22.2        &23.0       &16.7        &85.5         & 80.4        & 37.5        &28.3         &22.2         & 126.0        \\ 
+Weight-init     &77.3      &32.9        &27.0         &20.5       &111.0          & 80.6     & 37.9  & 28.6  &22.3     & 127.2         \\  \midrule
K=6: \\
\midrule
Base            &67.6       & 17.6         & 21.3     & 15.2      & 73.5        & 79.2        &32.2         &26.7         &20.4         &116.1         \\ 
+SeqKD          &73.2       &27.0        & 24.6      & 18.1       &  96.3       & 79.9        & 37.0        &28.0         & 21.6        & 123.9        \\ 
+Weight-init     &77.3      &33.3        &26.7         &20.4       &110.3          & 80.6     & 37.6  & 28.3   &22.2     & 126.2      \\ 
\bottomrule
\end{tabular}
    }
	\caption{The effect of sequence knowledge distillation~(SeqKD) and weight initialization~(Weight-init) . Beam width is set to $1$.  }
	\label{tricks}%
\end{table}

\begin{table}
	\centering
	\begin{tabular}{l|c|c|c|c|c}
		\toprule
		Model	&	b=1	& b=8 & b=16	& b=32	& b=64 \\
		\hline\hline
		Transformer  & 135ms& 22ms  & 13ms & 11ms & 10ms\\
		\hline
		SATIC, $K$=2	 & 76ms & 13ms  & 8ms & 7ms & 7ms\\
		SATIC, $K$=4	 & 46ms & 8ms  & 6ms & 5ms & 5ms\\
		SATIC, $K$=6   & 35ms & 7ms  & 5ms & 5ms & 5ms\\
		\bottomrule
	\end{tabular}
	\caption{Time needed to decode one sentence under various batch size settings. Beam width is set to $1$ since we find that larger beam width brings little performance boost but significant latency to SATIC model after self-critical training.}
	\label{latency}
\end{table}

\subsection{Qualitative Results}
\begin{figure}[!t] 
	\centering
	\includegraphics[width=3.3in]{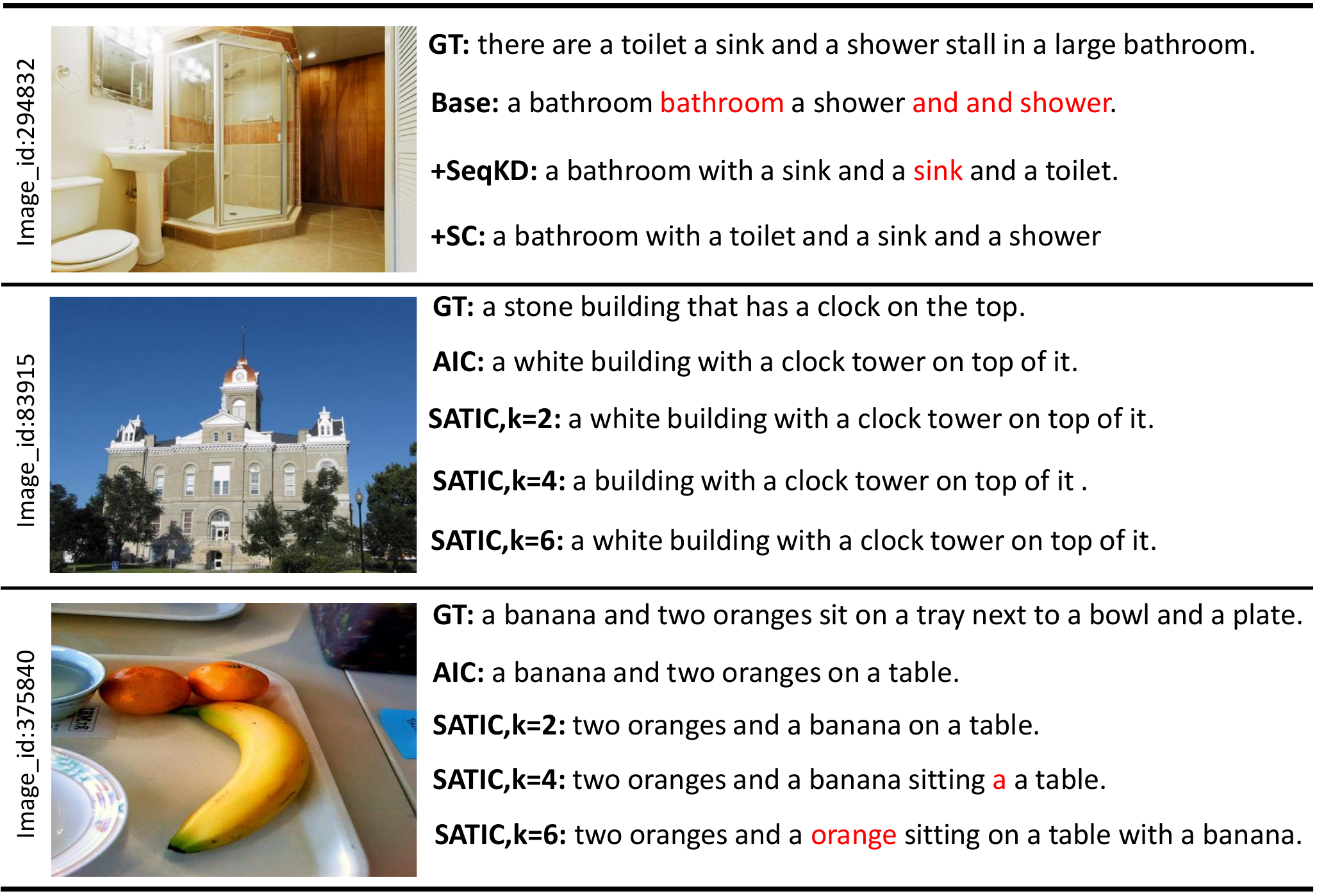}
	\caption{
		 Examples of the generated captions. 
		GT denotes  ground-truth caption. Base here denotes SATIC(k=4) model trained by  cross entropy loss using original training set. 
		We mark repeated words by red font.
	}
	\label{example}
\end{figure}

We present three examples of generated image captions in Figure~\ref{example}. 
From the top example, we can intuitively understand the effect of sequence knowledge distillation~(SeqKD) and self-crical training~(SC) in reducing repeated words and incomplete content. In general, the final SATIC models with different  group size $K$ can generate fluent captions, as shown in the middle example. Nevertheless, repeated words and incomplete content issues still exist, especially when $K$ is large. As shown in the bottom example, `\textit{a}' and `\textit{orange}' are repeated words and `\textit{tray}',`\textit{bowl}' are missing. 

\section{Conclusion} 
In this paper,  we  introduce  a  semi-autoregressive model  for  image  captioning  (dubbed  as  SATIC), which keeps the autoregressive property in global and non-autoregressive property in local. We conduct substantial experiments on  MSCOCO  image  captioning benchmark  to better understand the effect of each component. Overall, SATIC  achieves a better trade-off between speed,quality and easy training. 
\bibliographystyle{named}
\bibliography{SATIC}

\end{document}